\documentclass[conference]{IEEEtran}
\usepackage[utf8]{inputenc}       
\usepackage[a4paper,margin=2.5cm]{geometry}  

\usepackage{graphicx}     
\usepackage{caption}      
\usepackage{subcaption}   

\usepackage{float}        

\usepackage{enumitem}     

\usepackage[utf8]{inputenc}
\usepackage[a4paper,margin=2.5cm]{geometry}

\usepackage{graphicx}
\usepackage{float}        
\usepackage{dblfloatfix}  
\graphicspath{{image/}}
\usepackage{placeins}
\usepackage[utf8]{inputenc}
\usepackage[a4paper,margin=2.5cm]{geometry}

\usepackage{graphicx}
\usepackage{caption}    
\usepackage{cuted}      
\usepackage{placeins}
\usepackage{placeins,dblfloatfix}
\usepackage{amsmath}   
\usepackage{amssymb}   
\usepackage[utf8]{inputenc}
\usepackage{multicol}
\usepackage[square,authoryear]{natbib}   
\usepackage{url}
\usepackage{natbib}
\usepackage{multicol}

\ifCLASSINFOpdf
\else
\fi
\author{Pengyang Yu, Haoquan Wang, Gerard Marks,\\Tahar Kechadi , Laurence T. Yang, Sahraoui Dhelim and Nyothiri Aung}

\begin{document}
%
\title{MedLiteNet: Lightweight Hybrid Medical Image Segmentation Model}



%


\maketitle

\begin{abstract}
Accurate skin‐lesion segmentation remains a key technical challenge for computer-aided diagnosis of skin cancer.  
Convolutional neural networks, while effective, are constrained by limited receptive fields and thus struggle to model long-range dependencies.  
Vision Transformers capture global context, yet their quadratic complexity and large parameter budgets hinder use on the small-sample medical datasets common in dermatology.  We introduce the MedLiteNet, a lightweight CNN–Transformer hybrid tailored for dermoscopic segmentation that achieves high precision through hierarchical feature extraction and multi-scale context aggregation.  
The encoder stacks depth-wise Mobile Inverted Bottleneck blocks to curb computation, inserts a bottleneck-level cross-scale token-mixing unit to exchange information between resolutions, and embeds a boundary-aware self-attention module to sharpen lesion contours.  
On the ISIC 2018 benchmark, a single MedLiteNet model attains a Dice score of $0.897 \pm 0.010$ and an IoU of $0.821 \pm 0.015$ with fewer than $3.3\,\mathrm{M}$ parameters.  
A performance-weighted ensemble of three complementary variants raises accuracy to $0.904 \pm 0.012$ Dice and $0.830 \pm 0.018$ IoU while keeping the total parameter count below $10\,\mathrm{M}$—over $90\%$ smaller than Vision-Transformer backbones.  Qualitative results confirm superiority on irregular borders, low-contrast regions and multi-scale lesions, indicating MedLiteNet’s suitability for real-time, resource-aware computer-aided dermatology.
\end{abstract}


%
\IEEEpeerreviewmaketitle

\section{Introduction}
\subsection{Research background and challenges}
Skin cancer (especially melanoma) is a serious disease that threatens human health worldwide. Accurate automatic segmentation of skin lesion areas is of great significance for computer-aided diagnosis and treatment planning. It can help doctors more clearly assess the boundaries and areas of lesions, thereby improving the accuracy of early diagnosis and treatment effects. 

However, skin lesion segmentation faces many challenges: the color and texture of lesions in skin images vary widely, and often have low contrast with the surrounding normal skin; the shape and size of lesions vary significantly under different patients and imaging conditions, and the boundaries may be blurred. Traditional image segmentation algorithms (such as threshold methods, active contour models, etc.) require a lot of manual parameter adjustment and are difficult to cope with the complex and changeable morphology of skin lesions. 
\begin{figure}[H]
    \centering

    \begin{subfigure}{0.3\textwidth}
        \centering
        \includegraphics[width=\linewidth]{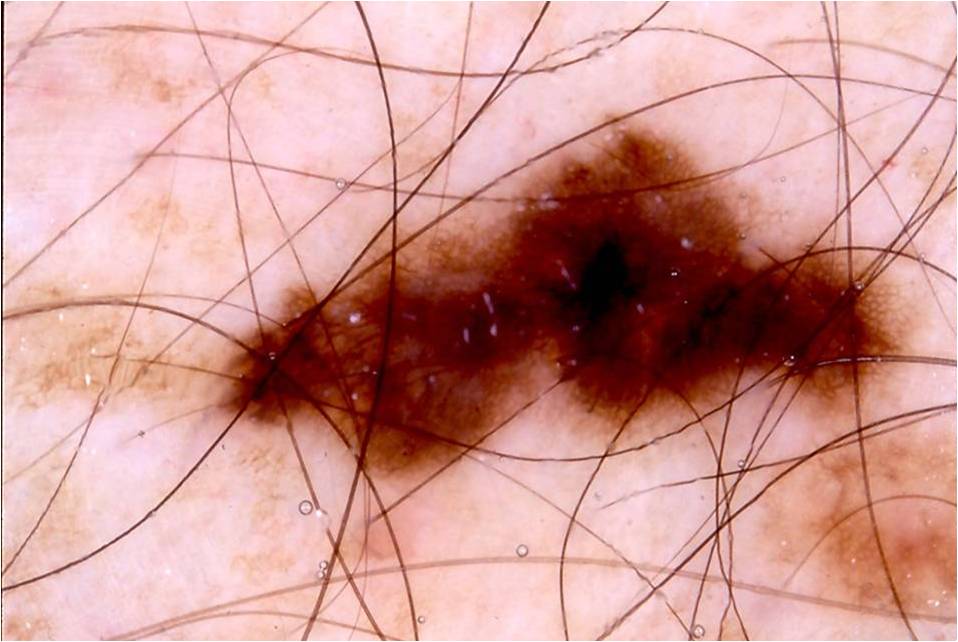}
        \caption{Original dermoscopic image}
    \end{subfigure}

    \vspace{0.2em} 

    \begin{subfigure}{0.3\textwidth}
        \centering
        \includegraphics[width=\linewidth]{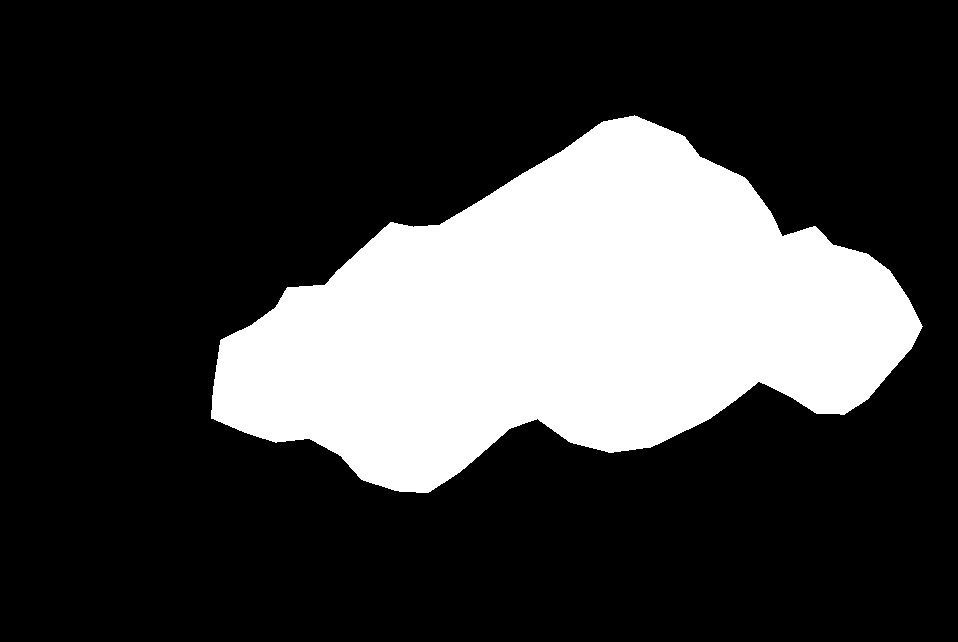}
        \caption{Lesion segmentation result}
    \end{subfigure}

    \caption{Comparison of original image and segmentation result for ISIC\_0000214.}
    \label{fig:segmentation-comparison}
\end{figure}
Therefore, in recent years, researchers have turned their attention to data-driven deep learning methods, hoping to use their powerful feature learning capabilities to solve the problem of skin cancer segmentation.

\subsection{Difficulties of existing technologies}

Traditional threshold, region growing and active contour methods rely on fixed priors and are difficult to adapt to images with blurred lesion edges and uneven brightness under different patients, imaging devices and lighting conditions. CNN, represented by U-Net, significantly improves segmentation accuracy by fusing multi-scale features through an encoding-decoding structure, but the local receptive field is limited and it is difficult to capture long-distance dependencies and large-scale deformations. Vision Transformer uses self-attention to achieve global modeling, but lacks inductive biases such as translation invariance. It is easy to underfit in small sample medical scenarios, and the self-attention calculation grows quadratically with the resolution, the reasoning is time-consuming and the edge characterization is insufficient.

Although hybrid architectures in recent years (such as TransUNet, Swin-Unet, TransFuse, FAT-Net) have both local and global advantages, they mostly rely on heavy backbones such as ResNet-50 or ViT-B, and the number of parameters and computing power overhead are tens of megabytes, which is not conducive to mobile or real-time clinical deployment. Therefore, how to effectively compress the model size and reduce the reasoning delay while maintaining segmentation accuracy has become a key issue for medical image segmentation to move from the laboratory to real applications.

\subsection{Our Method and Contributions }
MedLiteNet proposes a lightweight CNN-Transformer hybrid architecture for ISIC 2018 dermoscopy images, with MobileNetV2 inverted residual as the backbone. The cross-scale Local-Global Block fuses the local texture of the depthwise separable convolution and the global dependencies captured by the self-attention mechanism, and injects boundary-aware weights into the attention mechanism to enhance blurred contours.

The decoder uses SCSE recalibration and enhanced ASPP in parallel, and multi-scale aggregation maintains good localization accuracy while maintaining fine localization. The model has only about 3 million parameters. After size-increment training optimization, the single-frame inference time for 256 × 256 images is about 1 millisecond (RTX A6000), with extremely low computational and storage overhead.

In the ISIC 2018 evaluation, the final integrated test achieved a validation set Dice of 0.913/IoU 0.84 and a test set Dice of 0.905/0.83, achieving lightweight while maintaining segmentation performance that is almost the same as the best method.

The experiment shows that the carefully designed lightweight hybrid architecture can maintain high-precision skin lesion segmentation at extremely low computational complexity, meeting the deployment requirements of resource-constrained scenarios (such as mobile devices and primary healthcare), and is expected to promote the popularization of early skin cancer screening technology.

Our contributions are summarized as follow:
\begin{itemize}
  \item We propose a “local--global--boundary’’ triple-coupled architecture that delivers clinical-grade accuracy with a mobile-scale footprint.%
  \item We developed a boundary-aware self-attention mechanism that explicitly reinforces contour representation.%
  \item We propose a progressive size-increment training and FP16 inference recipe that processes a $256\times256$ image in $1\,\mathrm{ms}$ raw latency ($23\,\mathrm{ms}$ with six-fold TTA).
  \item Extensive experiments showing that carefully designed lightweight networks, even when ensembled, can rival heavy models in realistic dermatology scenarios.
\end{itemize}

\section{Related Works }
\subsubsection{CNN-based Segmentation Methods }
CNN has become the mainstream technology for medical image segmentation with its excellent local feature extraction capabilities. Among them, the U-Net model with an encoder-decoder structure is the most representative. U-Net extracts high-level semantic features by downsampling and fuses high-resolution details through skip connections, achieving good segmentation of fine structures in biomedical images. Various improved models of U-Net have also emerged. For example, the U-Net++ model introduces dense skip connections between encoder layers, further optimizing feature fusion through nested structures; Attention U-Net introduces an attention gate mechanism at the skip connection to suppress irrelevant regional features to highlight the target area. These CNN models have achieved excellent results on many public datasets. 

However, the inherent limitation of CNN is that the receptive field is limited and it mainly focuses on local neighborhood pixels. Although the global field of view can be expanded by deepening the network or dilated convolution, it is still difficult to effectively capture long-distance dependencies. In addition, CNN usually relies on downsampling for feature compression, which may lead to the loss of fine-grained spatial information. For skin lesions with blurred boundaries and small sizes, pure CNN models are often unable to cope with it.

\begin{figure}[htbp]
  \centering
  \begin{subfigure}[t]{0.32\textwidth}
    \centering
    \includegraphics[width=\linewidth]{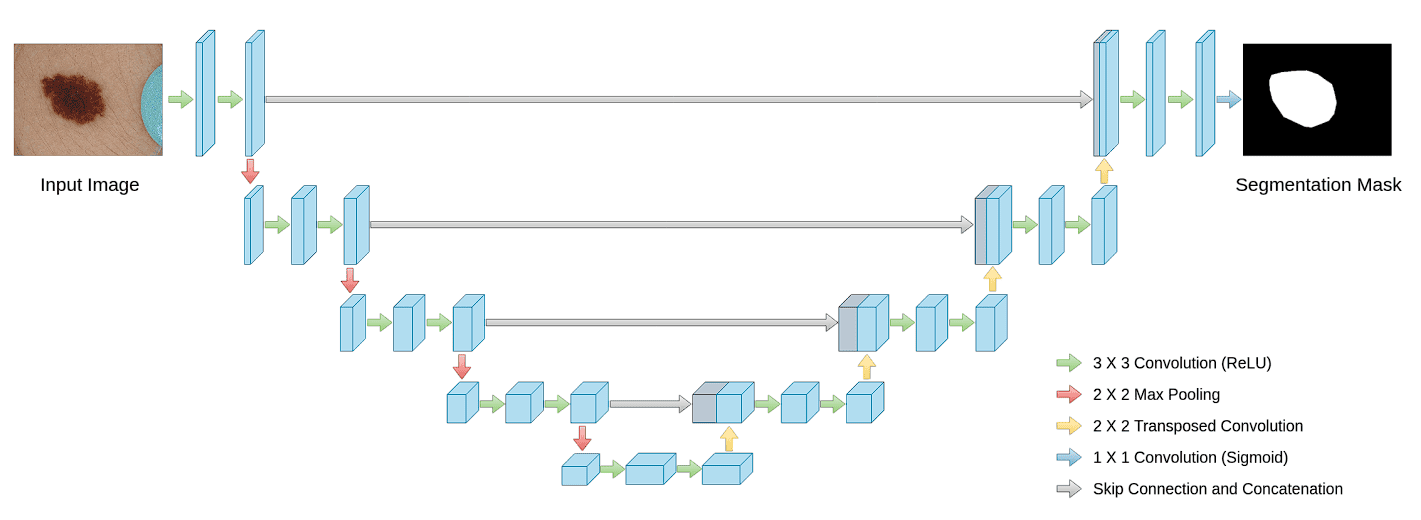}
    \caption{UNet Structure}
    \label{fig:unet}
  \end{subfigure}\hfill
  \begin{subfigure}[t]{0.32\textwidth}
    \centering
    \includegraphics[width=\linewidth]{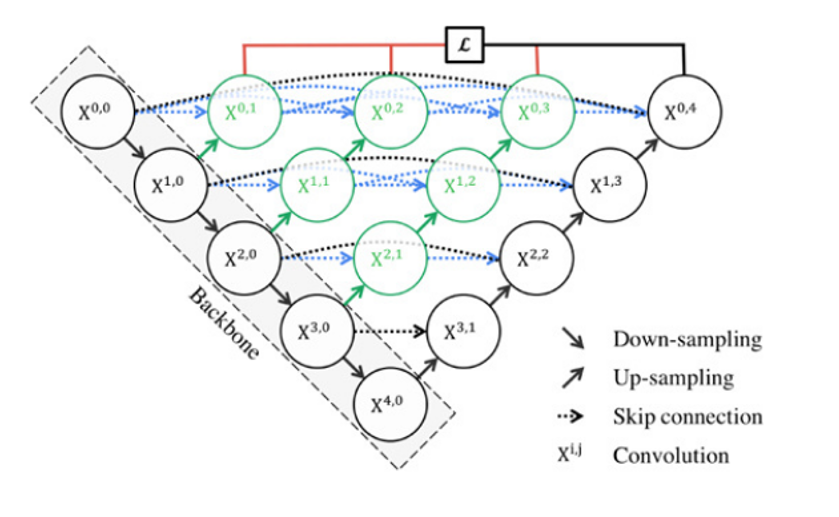}
    \caption{UNet++ Structure}
    \label{fig:unetpp}
  \end{subfigure}\hfill
  \begin{subfigure}[t]{0.32\textwidth}
    \centering
    \includegraphics[width=\linewidth]{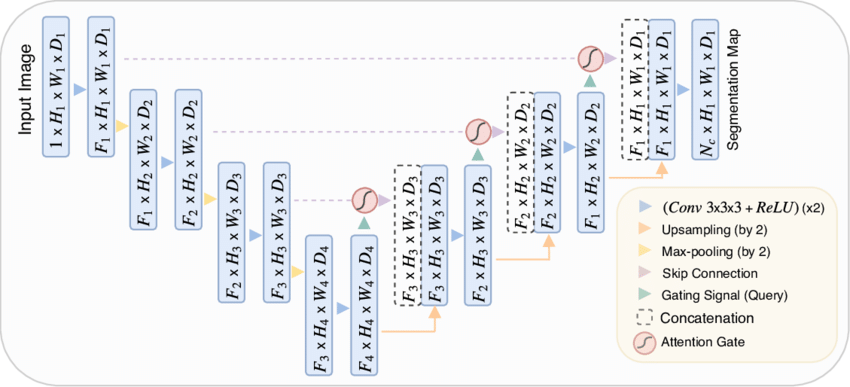}
    \caption{AttentionUNet Structure}
    \label{fig:attunet}
  \end{subfigure}
  \caption{Comparison of three common convolutional segmentation network structures}
  \label{fig:cnn-comparison}
\end{figure}

\subsubsection{Transformer-based Segmentation Methods }
The introduction of the Transformer architecture provides a new opportunity for global information modeling. Transformer models the correlation between any two positions in a sequence through the self-attention mechanism, and has made breakthroughs in tasks such as natural image classification. Vision Transformer (ViT) divides images into patch embedding sequences, and for the first time proves the feasibility of pure Transformer in visual tasks. 

In the field of medical image segmentation, the TransUNet model proposed by Chen et al. embeds the pre-trained ViT into the U-Net encoder, uses the Transformer to encode long-range dependencies, and uses the features extracted by CNN to retain local details. TransUNet significantly improves the accuracy in tasks such as multi-organ segmentation. Subsequently, models such as Swin-Unet use a hierarchical window attention mechanism to improve the efficiency of the Transformer, which can reduce computational overhead while maintaining global modeling. 

\begin{figure}[htbp]
    \centering
    \includegraphics[width=0.53\textwidth]{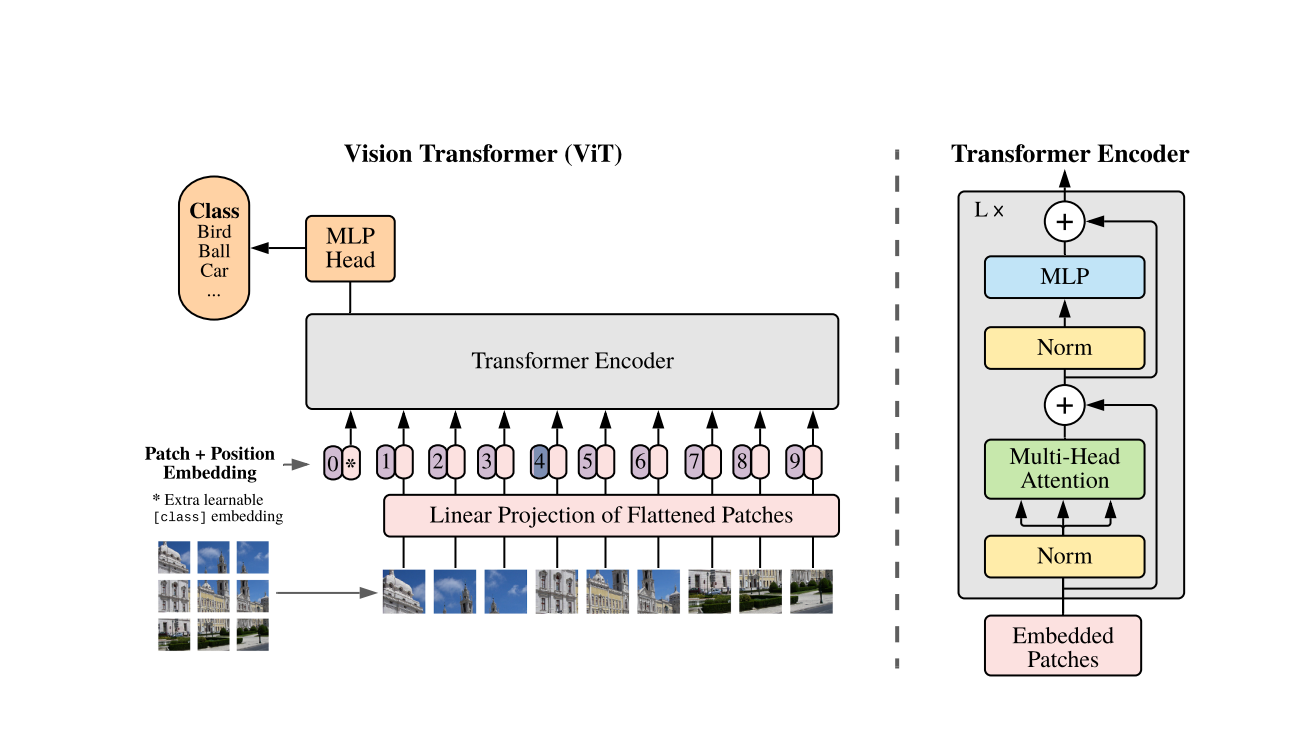}
    \caption{Vision Transformer structure for image-level representation learning via patch embeddings and self-attention. }
    \label{fig:vit}
\end{figure}

Although the Transformer has expanded the receptive field, it still has some problems in the application of medical image segmentation: on the one hand, the Transformer model has a huge number of parameters and requires massive data pre-training. The scale of medical data sets such as ISIC is relatively limited, and direct training of the Transformer is prone to overfitting; on the other hand, forcibly embedding the Transformer module into the CNN architecture may bring about differences in feature distribution, resulting in inconsistent fusion, thereby affecting segmentation performance. In addition, ViT's method of dividing images into patches will lose some spatial details, which is also a major obstacle for segmentation tasks that require pixel-level fine positioning.

\subsubsection{CNN-Transformer Hybrid Architecture }

To reconcile the strength of CNNs in encoding local details with the Transformer's ability to capture long-range context, a growing number of studies have adopted hybrid designs. Early attempts, such as TransUNet (2021), grafted a pre-trained ViT encoder onto a U-Net backbone, achieving 88.5 \% Dice / 83.7 \% IoU at the cost of 105 M parameters. 

Subsequent efforts reduced complexity: FAT-Net (2022) combined a lightweight CNN with a windowed self-attention mechanism, cutting parameters to 28 M while slightly improving the Dice to 89.0 \%. More recent methods have further pushed the efficiency boundary—BACANet (2024) introduced boundary-aware convolutions and attention gating, reaching a state-of-the-art 92.1 \% Dice / 85.4 \% IoU with only 7.56 M parameters. However, even this footprint may be prohibitive for mobile or point-of-care devices. 

Our proposed \textit{MedLiteNet} (2025) compresses the hybrid paradigm to just 3.2 M parameters—more than 98 \% less than TransUNet—while maintaining competitive accuracy (Dice 90.5 \%, IoU 83.0 \%). The comparison in Table~\ref{tab:hybrid_review} shows that model size has fallen by two orders of magnitude in four years, yet a truly “Pareto-optimal” balance between accuracy and efficiency remains elusive.

\begin{table}[H]
\centering

\renewcommand{\arraystretch}{1.2}
\setlength{\tabcolsep}{2pt}
\begin{tabular}{|c|c|c|c|c|}
\hline
\textbf{Model} & \textbf{Year} & \textbf{Params} & \textbf{Dice (\%)} & \textbf{IoU (\%)} \\
\hline
TransUNet& 2021 & 105M& 88.5 & 83.7\\\hline

FAT-Net& 2022 & 28M& 89.0 & 80.2\\\hline

BACANet& 2024 & 7.56M & 92.1 & 85.4\\\hline

MedLiteNet & 2025 & 3.2M& 90.5& 83.0\\
\hline
\end{tabular}
\caption{Summary of CNN–Transformer hybrid models.}
\label{tab:hybrid_review}
\end{table}

MedLiteNet aims to bridge this gap through:
\begin{itemize}
    \item Depth-based MBConv encoder – leverages depth-wise separable Mobile Inverted Bottleneck (MBConv) blocks to build a compact backbone while preserving local texture representation;
    \item Cross-scale token hybrid module – fuses convolutional features with cross-scale tokens, using self-attention to capture long-range context;
    \item ASPP + SCSE decoder – combines Atrous Spatial Pyramid Pooling (ASPP) with parallel Spatial-Channel Squeeze-and-Excitation (SCSE) attention for multi-scale semantics and fine-grained boundary recovery.
\end{itemize}

This lightweight yet high-performance configuration makes \textit{MedLiteNet} feasible for deployment in resource-constrained clinical settings.

\subsubsection{Differences from this paper’s method}
Unlike existing hybrid architectures that typically adopt heavy backbones such as ResNet-50 or ViT-B and apply Transformers to high-dimensional feature sequences, the proposed MedLiteNet employs a Mobile Inverted Bottleneck as its backbone, introduces a cross-scale Local-Global Block at the encoding bottleneck, and leverages depthwise separable convolutions to preserve local textures. Meanwhile, lightweight self-attention is applied to model global dependencies, with boundary-aware weights injected into the attention mechanism. Additionally, ASPP and SCSE modules are integrated at the decoder to enhance multi-scale semantics and contour details.

Owing to this lightweight "local-global-boundary" triple coupling design, MedLiteNet contains only 3.2M parameters, achieving significantly faster inference compared to models like TransUNet and Swin-Unet (typically 40--100M). Despite its compactness, MedLiteNet combines model performance and parameters to achieve an excellent Dice score of $\approx 0.91$ on the ISIC 2018 dataset, demonstrating its high efficiency and accuracy in skin lesion segmentation.

\section{Methodology}
\subsection{Overall Model Architecture}
The lightweight CNN-Transformer hybrid model proposed in this paper adopts the overall architecture of encoder-decoder, as shown in Figure 4.

\begin{figure*}[!htb]
  \centering
  
  \includegraphics[width=\textwidth]{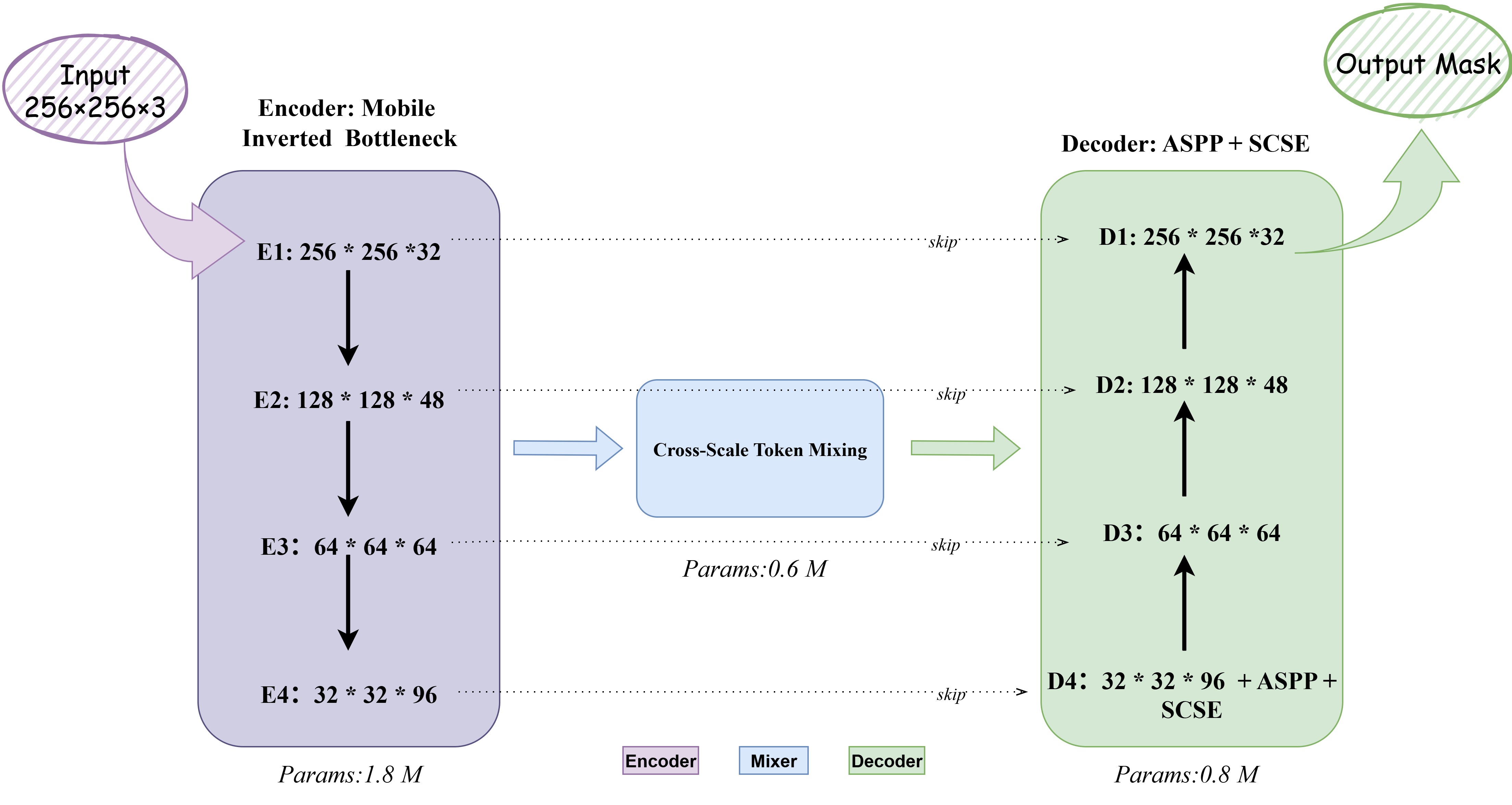}
  \caption{Overall architecture of the proposed MedLiteNet, which consists of a lightweight encoder, a cross‐scale token mixer, and a decoder equipped with ASPP and SCSE modules.}
  \label{fig:medlitenet-arch}
\end{figure*}

The encoder of the model consists of a convolution branch and a Transformer branch, which is used to extract multi-scale feature representations of the input skin lesion image; the decoder fuses the features of different scales from the encoder step by step and upsamples them to generate a binary segmentation mask of the same size as the input. The encoder first uses a series of lightweight convolution modules to extract local features and embeds the boundary-aware attention mechanism to enhance the edge information of the lesion; then, a local-global fusion block is introduced to process local convolution features and global self-attention features in parallel to achieve local-global information interaction. At the end of the encoding, the ASPP pyramid pooling module is used to further extract multi-scale context features to enhance the recognition ability of lesions of different sizes. The decoder part is similar to the classic U-Net. The features of each stage of the encoder are transferred to the corresponding decoding layer through jump connections, the low-level details and high-level semantics are fused, and the segmentation map is gradually restored through upsampling and convolution.

In particular, We also post-process,techniques during decoding to pay more attention to the blurred boundary,regions when reconstructing the segmentation mask. The model finally outputs a two-dimensional probability map of the same size as the input image, and obtains the binary segmentation result after threshold processing.

\subsection{Lightweight Convolutional Encoder (MBConv)}

To minimise model size and computation while preserving segmentation
accuracy, we build the encoder exclusively from MobileNet-V2’s
\textit{inverted residual} block (MBConv).  
An MBConv first applies a  \(1\times1\)  point-wise convolution that expands
the feature map from  \(C\)  channels to  \(C_{\mathrm{e}}=tC\) ,
with expansion factor  \(t=6\) .
It then performs a \(3\times3\) \emph{depthwise} convolution to extract
spatial features channel-wisely, and finally projects the tensor back to
 \(C^{\prime}\)  channels using another  \(1\times1\)  convolution before
residual addition.  
This “inverted” design keeps the widest layer in the middle, so the
depthwise stage retains sufficient representational capacity.

\vspace{2pt}
\noindent
Whereas a conventional residual block requires
 \begin{equation}
  K^{2}C_{\text{in}}C_{\text{out}}
\end{equation}
weights,
an MBConv decomposes the cost into
\begin{equation}
  K^{2}C_{\text{in}} \;+\; C_{\text{in}}C_{\text{out}},
\end{equation}
achieving nearly an order-of-magnitude reduction in both parameters and
multiply–adds when \(K{=}3\) and \(t{=}6\).

\vspace{2pt}
\noindent
Given a  \(3\times256\times256\)  dermoscopic image, the encoder first
employs a stride-2  \(3\times3\)  convolution for coarse down-sampling,
followed by four MBConv stages with channel widths
 \([32,\,64,\,128,\,256]\).
The first block of each stage uses stride-2 for additional reduction,
while the remaining blocks keep stride-1 for feature refinement;
all convolutions are followed by \textit{BatchNorm} and \textit{SiLU}.  
This layout yields only  \(\sim\!1.8\text{M}\)  parameters—two orders of
magnitude fewer than the tens of millions typical in a U-Net
encoder—and proportionally shortens inference latency, leaving ample
computational budget for the subsequent Transformer module.

\subsection{Transformer global encoding module and local-global fusion}

To compensate for the limited global context modeling of convolutional networks, we introduce a Transformer‐based global encoding module at the encoder bottleneck. Given a convolutional feature map
\begin{equation}
F \in \mathbb{R}^{H \times W \times C},
\end{equation}
we first flatten it into a sequence of \(N = H \times W\) tokens of dimension \(C\). Each token is projected via a learnable linear layer into a \(d\)-dimensional embedding, and a positional encoding is added to preserve spatial information. We then apply \(L\) stacked Transformer layers—each consisting of a Multi‐Head Self‐Attention (MHSA) sublayer and a feed-forward network (FFN) sublayer, both wrapped with residual connections and LayerNorm. The MHSA computes pairwise dot-product attention across all tokens, capturing long-range dependencies that relate lesion regions to the entire skin context. This yields a global feature sequence
\begin{equation}
G \in \mathbb{R}^{N \times d},
\end{equation}
which we reshape back into a 2D feature map of size  \(\tfrac{H}{r} \times \tfrac{W}{r} \times d\), where \(r\) is the overall downsampling factor.

\medskip

At the same bottleneck, we fuse these global Transformer features with the corresponding convolutional features \(F_{\mathrm{conv}}\). First, both feature maps are aligned to the same spatial resolution \(\tfrac{H}{r}\times \tfrac{W}{r}\) (using average pooling or strided convolution on \(F_{\mathrm{conv}}\) if necessary) and to the same channel dimension \(d\). We then perform pixel-wise fusion by concatenating along the channel axis and applying a \(1\times 1\) convolution, followed by a residual addition:
\begin{equation}
F_{lg}
\;=\;
\sigma\bigl(W_{1} * [\,F_{\mathrm{conv}}\;\Vert\;F_{\mathrm{trans}}\,] + b_{1}\bigr)
\;+\;
F_{\mathrm{conv}}
\;+\;
F_{\mathrm{trans}},
\end{equation}
where \([\Vert]\) denotes channel‐wise concatenation, \(W_{1}, b_{1}\) are the weights and bias of the \(1\times1\) convolution, and \(\sigma\) is a non-linear activation such as ReLU. This strategy merges the fine‐detail sensitivity of convolutional features with the holistic, long‐range context captured by the Transformer, producing a semantically rich representation that is then passed to the subsequent ASPP module and decoder.

It should be emphasized that this work chooses to perform local-global fusion at the bottleneck layer for the sake of balancing efficiency and effect. On the one hand, the size of the bottleneck layer feature map is relatively small, and the Transformer can process it at a low cost; on the other hand, the features already contain information at a higher semantic level, and fusing global dependencies can maximize the semantic discrimination required for segmentation decisions. If the Transformer is introduced at a higher resolution layer, the computational overhead will increase exponentially, while introducing it at a lower semantic layer will only have limited improvement on the results.

\subsection{Boundary‐Aware Attention Module}
Precise segmentation of skin lesions relies on accurate boundary detection. However, lesion edges are often blurred or exhibit colors similar to surrounding skin, causing standard convolutional or Transformer methods to miss or misclassify boundary pixels. To address this, we propose a Boundary‐Aware Attention (BAA) module that enhances the focus on edge pixels during feature extraction, improving the mask’s boundary alignment.

Module design: The BAA module is inspired by attention‐gate mechanisms and consists of three main stages:

\begin{itemize}
  \item Boundary Feature Extraction:  
    Apply a $3\times3$ convolution with Laplacian‐like filters or compute spatial gradients to obtain a boundary response map
   \begin{equation}
      B \in \mathbb{R}^{H\times W}.
  \end{equation}
    Optionally, incorporate global context from Transformer features $F_{\mathrm{trans}}$ via a linear projection and addition:
    \begin{equation}
      B' = B + W_{g}\,\mathrm{flatten}(F_{\mathrm{trans}}),
    \end{equation}
    where $W_{g}$ is a learnable weight matrix.

  \item Attention Weight Computation:  
    Concatenate the boundary map $B'$ with the original feature map $F\in\mathbb{R}^{H\times W\times C}$ along the channel axis, then apply a $1\times1$ convolution followed by a Sigmoid activation to produce the attention mask
    \begin{equation}
      M = \sigma\bigl(\mathrm{Conv}_{1\times1}([F \,\Vert\, B'])\bigr)
      \in [0,1]^{H\times W}.
    \end{equation}

  \item Feature Refinement:  
    Re‐weight the original features by emphasizing boundary regions:
    \begin{equation}
      F' = F \;\otimes\; \bigl(1 + M\bigr),
    \end{equation}
    where $\otimes$ denotes element‐wise multiplication, amplifying boundary pixel features while keeping non‐boundary features largely unchanged.
\end{itemize}

We insert the BAA module at two locations: at the end of the encoder—where high‐level semantics guide boundary prediction with reduced noise—and at the final stage of the decoder—to correct any boundary shifts introduced by up‐sampling. This streamlined design leverages existing feature maps to infer boundary attention, enhancing mask fidelity for clinical measurements of lesion diameter and area.

\subsection{ASPP Multi-Scale Context Module}
Skin lesion morphology varies greatly—from a few pixels to large contiguous regions—requiring the model to possess multi‐scale detection capability. Although the encoder’s downsampling imparts a degree of scale invariance, explicit fusion of multi‐scale features is still necessary for robustness. Therefore, we introduce an Atrous Spatial Pyramid Pooling (ASPP) module immediately after the local–global feature fusion block.

ASPP extracts multi‐scale features by applying parallel dilated convolutions with different dilation rates. We employ four $3\times3$ dilated convolutions with dilation rates $r_{1}=1$, $r_{2}$, $r_{3}$, and $r_{4}=12$, alongside a global average pooling branch. Low dilation rates capture local details, while high dilation rates expand the receptive field to capture global context. Each branch output is passed through a $1\times1$ convolution to adjust the channel dimension, then upsampled to the original resolution, concatenated along the channel axis, and finally fused via a $1\times1$ convolution to produce the multi‐scale enhanced feature map F\_aspp.

This design enables the network to adaptively handle lesions of various sizes: small lesions are precisely localized by the low‐dilation branches, large lesions benefit from extended receptive fields to capture full contour information, and the global pooling branch provides image‐level background reference to suppress false positives such as illumination artifacts. Because ASPP operates only on the low‐resolution bottleneck feature map (with output channels constrained to 256–512), the additional computational overhead is modest. Empirical results show that integrating ASPP consistently improves Dice and IoU metrics, with especially pronounced gains for lesions at the extremes of the size spectrum.

\subsection{Decoder and output layer} 
The boundary-enhanced fused features $F_{enc}$ from the encoder, containing global, local, and multi-scale information, are fed into the decoder for upsampling reconstruction. The decoder adopts a symmetric U-Net architecture, leveraging skip connections to fuse multi-level features. Each decoding stage performs: (1) feature upsampling to the previous resolution level; (2) concatenation with corresponding encoder features followed by convolutional fusion. This design effectively preserves fine-grained details from shallow layers, compensating for information loss during upsampling.

In our implementation, the bottleneck features $F_{enc}$ pass through four decoding stages, with each stage fusing features from the corresponding encoder level. Except for the final stage, boundary attention modules are embedded in each decoding stage to further refine edge quality using high-resolution features. The final output is generated through a $1 \times 1$ convolution followed by sigmoid activation to produce a probability map, which is then thresholded at 0.5 to obtain the binary segmentation mask $\hat{Y}$.

Loss Function Design: We employ a weighted combination of Dice loss and Binary Cross-Entropy (BCE) loss to optimize both regional overlap and pixel-wise accuracy:

\begin{itemize}
    \item Dice Loss: Measures the overlap between predicted and ground truth masks:
    \begin{equation}
    L_{\text{Dice}} = 1 - \frac{2\sum_{i=1}^{n} p_i g_i + \varepsilon}{\sum_{i=1}^{n} p_i + \sum_{i=1}^{n} g_i + \varepsilon}
    \end{equation}
    where $p_i$ denotes the predicted probability, $g_i$ represents the ground truth label, and $\varepsilon$ is a smoothing term for numerical stability. The Dice loss is particularly sensitive to overall regional matching, making it suitable for handling class imbalance issues common in lesion segmentation.
    
    \item Binary Cross-Entropy (BCE) Loss: Optimizes pixel-wise classification accuracy:
    \begin{equation}
    L_{\text{BCE}} = -\frac{1}{n}\sum_{i=1}^{n} \left( g_i \log p_i + (1-g_i)\log(1-p_i) \right)
    \end{equation}

    This loss ensures accurate pixel-level predictions across the entire image.
    
    \item Combined Loss Function: The total loss is formulated as:
    \begin{equation}
    L_{\text{total}} = \alpha L_{\text{BCE}} + \beta L_{\text{Dice}}
    \end{equation}

    where we set $\alpha = \beta = 0.5$ in our experiments. This combination ensures both pixel-level classification performance and overall regional matching accuracy, effectively improving segmentation precision and training stability.
\end{itemize}

Experimental results demonstrate that this loss function design outperforms single loss functions or more complex schemes with additional boundary constraints.

\section{Data and Experiments}
\subsection{Datasets}
This study employs the ISIC 2018 skin lesion segmentation dataset, which comprises dermoscopic images with lesion regions professionally annotated by dermatologists. The dataset contains approximately 2,594 training images with resolutions ranging from 540×576 to 4,499×6,748 pixels. We adopt the official train/test split to ensure consistency and fairness in comparison with other methods. Given the high resolution of original images, we randomly crop and resize them to 512×512 pixels before feeding into the model. The segmentation task is formulated as a binary classification problem, where lesion pixels are labeled as foreground and healthy skin as background.

To enhance model robustness, we implement a comprehensive data augmentation strategy during training. Geometric transformations include random horizontal/vertical flipping and rotation to improve the model's invariance to lesion position and orientation variations. We further incorporate spatial transformations such as elastic deformation, scaling, and translation to simulate skin tissue deformation. For pixel intensity augmentation, we apply color and brightness perturbations including random brightness-contrast adjustments, CLAHE histogram equalization, and gamma value modifications to enhance the model's adaptability to illumination variations. Additionally, Gaussian noise and blur are randomly introduced to improve noise robustness. All images undergo normalization using ImageNet dataset statistics (mean and standard deviation) to eliminate pixel value distribution differences across images before being fed into the neural network.

\subsection{Evaluation Metrics}
We adopt standard medical image segmentation metrics, namely Dice coefficient and Intersection over Union (IoU), as our primary evaluation criteria:

\begin{itemize}
    \item Dice Coefficient:Measures the overlap between predicted and ground truth foreground regions, with values ranging from [0,1]. Higher values closer to 1 indicate better segmentation quality. A Dice score exceeding 0.9 is typically considered high-quality segmentation.
    
    \item IoU (Jaccard Index):Computes the ratio of intersection to union between predicted and ground truth foreground regions:
    \begin{equation}
    \text{IoU} = \frac{|P \cap G|}{|P \cup G|}
    \end{equation}
    where $P$ and $G$ denote the predicted and ground truth regions, respectively. IoU is mathematically related to Dice through:
    \begin{equation}
    \text{Dice} = \frac{2\text{IoU}}{\text{IoU}+1}
    \end{equation}
    This metric intuitively reflects the degree of overlap between predicted and actual lesion regions.
\end{itemize}

Although we also record auxiliary indicators such as pixel-level accuracy, sensitivity (recall rate) and specificity to comprehensively evaluate the performance, due to the serious imbalance of categories in the skin lesion segmentation task (background pixels dominate), the accuracy and specificity are often high due to the large proportion of background, which cannot fully reflect the model's ability to segment foreground lesions. Therefore, Dice and IoU, as more balanced and comprehensive evaluation indicators, will be used as the main performance measurement criteria for this study.

\subsection{Experimental Setup}
Our model is implemented using the PyTorch deep learning framework, with all experiments conducted on a single shared NVIDIA RTX A6000 GPU. The training configuration employs a batch size of 16 for 165 epochs, with an initial learning rate of $1 \times 10^{-3}$ and AdamW optimizer.

To ensure training stability and efficiency, we implement three key strategies:
\begin{itemize}
    \item Gradient clipping: Gradient norms are clipped at 0.5 to prevent gradient explosion and maintain stable convergence.
    \item Mixed-precision training: Automatic mixed precision (AMP) with FP16/FP32 is utilized to accelerate training and reduce memory consumption.
    \item Exponential moving average (EMA): Model weights are updated with a decay rate of 0.999 to enhance generalization performance and stability.
\end{itemize}

The learning rate schedule follows a cosine annealing strategy, smoothly decaying from the initial value to near zero according to a cosine function, ensuring fine-grained convergence in later training stages. Additionally, gradient accumulation (updating weights every 2 iterations) is employed to effectively simulate larger batch training.

Model selection is based on validation set performance using Dice coefficient and IoU metrics, with final evaluation conducted on an independent test set. Table 2 summarizes the key hyperparameter settings used in our experiments.
\begin{table}[t]  

  \centering
  \begin{tabular}{|@{}c|c@{}|}
    \hline
    \textbf{Hyper-parameter}             & \textbf{Setting}            \\ \hline
    Deep-learning framework              & PyTorch                     \\\hline
    GPU device                           & NVIDIA RTX\,A6000           \\\hline
    Batch size                           & 16                          \\\hline
    Training epochs                      & 300                         \\\hline
    Initial learning rate                & $1 \times 10^{-3}$          \\\hline
    Optimizer                            & AdamW                       \\\hline
    LR scheduler                         & Cosine annealing            \\\hline
    Gradient clipping                    & Yes (max-norm $=0.5$)       \\\hline
    Mixed-precision training             & Yes (FP16 AMP)              \\\hline
    EMA weight averaging                 & Yes (decay $= 0.999$)       \\ \hline
  \end{tabular}
\caption{Training hyper-parameters for MedLiteNet}
\label{tab:hyper}
\end{table}

\section{Experimental Results and Visualization}

\subsection{Quantitative Results Comparison}
As shown in Table~\ref{tab:hybrid_review}, the proposed MedLiteNet achieves comparable but more lightweight performance to the current state-of-the-art models (e.g., TransUNet, FAT-Net, BACANet) on all skin lesion comprehensive segmentation evaluation metrics. In terms of key metrics - Dice coefficient and intersection over union (IoU), MedLiteNet shows comparable segmentation accuracy to these methods, with some metrics showing marginal improvements.

It is worth noting that MedLiteNet contains only about three million parameters, making it the most lightweight model among all compared methods. This represents a significant reduction compared to models like TransUNet, which typically contain tens or even hundreds of millions of parameters. Despite this remarkably compact model size, MedLiteNet maintains competitive segmentation accuracy. This demonstrates that our proposed model successfully achieves an optimal balance between efficiency and effectiveness, substantially reducing model complexity while preserving high segmentation performance.

\subsection{Segmentation result visualization}

\begin{figure}[H]
    \centering
    \includegraphics[width=\columnwidth]{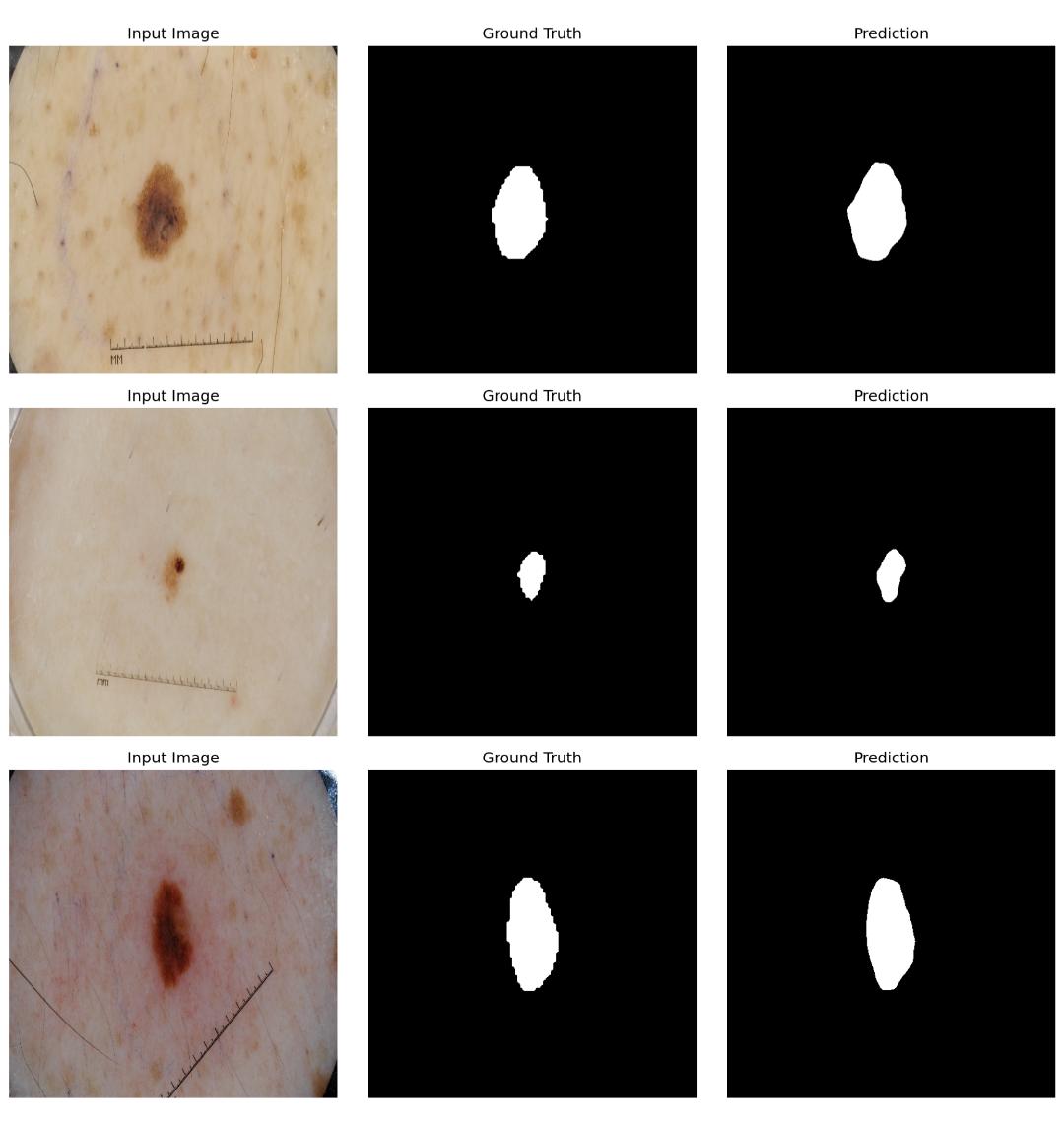}
    \caption{Qualitative segmentation results on ISIC 2018 dataset. Each row shows (from left to right): input dermoscopic image, ground truth annotation, and MedLiteNet prediction. The results demonstrate accurate lesion boundary delineation across diverse lesion morphologies.}
    \label{fig:qualitative}
\end{figure}

Figure~\ref{fig:qualitative} presents qualitative segmentation results, displaying dermoscopic input images alongside ground truth (GT) annotations and MedLiteNet predictions. Visual inspection reveals high consistency between predicted masks and ground truth in terms of both shape and boundary delineation.

For lesions with regular morphology, MedLiteNet accurately captures the lesion contours, with predicted masks exhibiting near-perfect overlap with ground truth annotations. The model preserves fine boundary details, demonstrating robust lesion recognition capability, particularly for cases with well-defined boundaries.

However, challenges remain in complex scenarios. For lesions with ambiguous boundaries or irregular shapes, the model occasionally exhibits minor under- or over-segmentation at the periphery. Additionally, surface artifacts such as hair follicles or ruler markings are sporadically misclassified as lesions, resulting in isolated false positives. These observations provide valuable insights for future model refinement.

Overall, the qualitative analysis corroborates our quantitative findings, confirming MedLiteNet's effectiveness in skin lesion segmentation. The visual results demonstrate that the model achieves expert-level annotation quality in the majority of cases, nearly validating its clinical applicability.

\subsection{Training Convergence Analysis }

\begin{figure}[htbp]
  \centering
  
  \includegraphics[width=0.55\columnwidth]{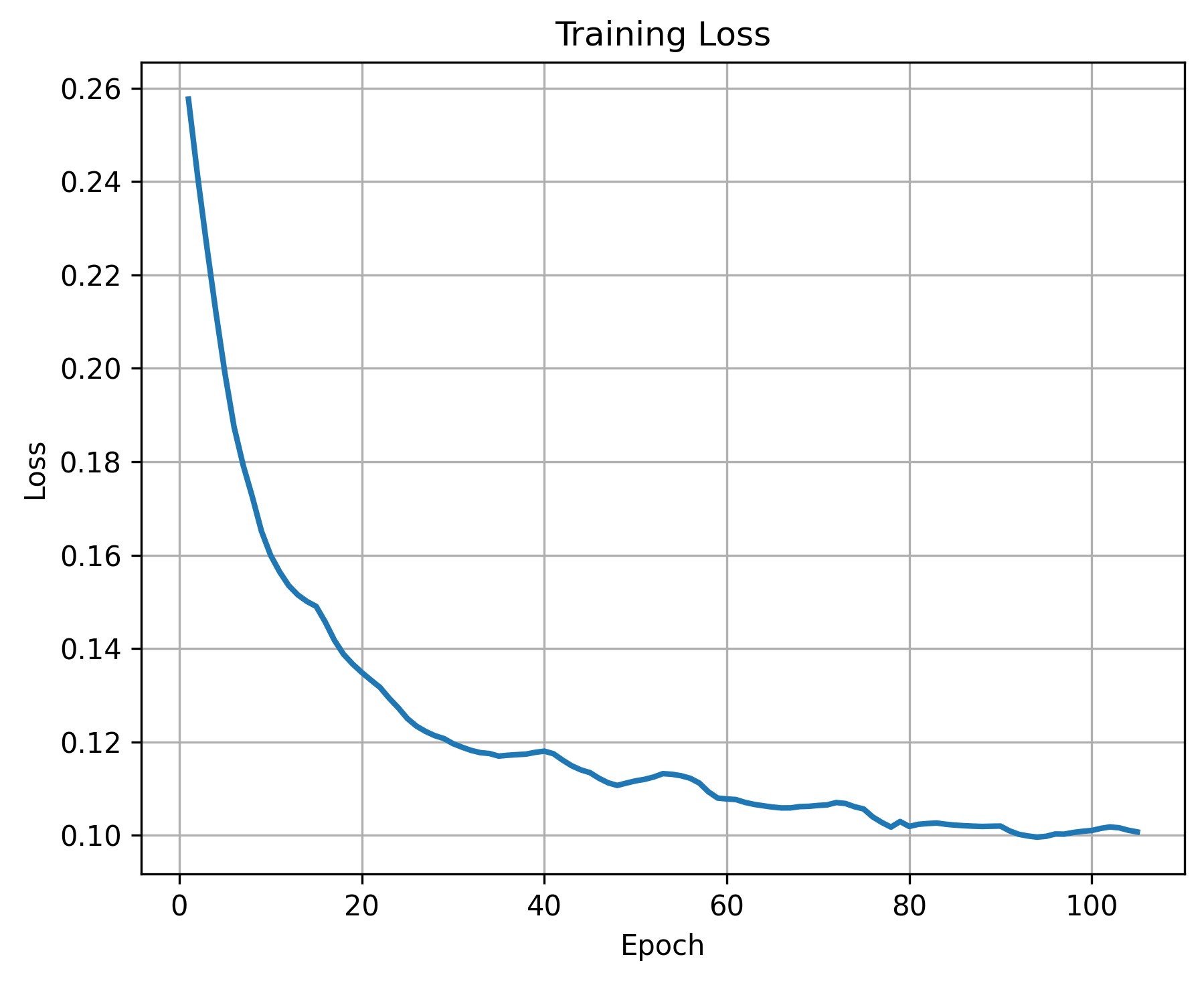}\\[1ex]
  
  \includegraphics[width=0.55\columnwidth]{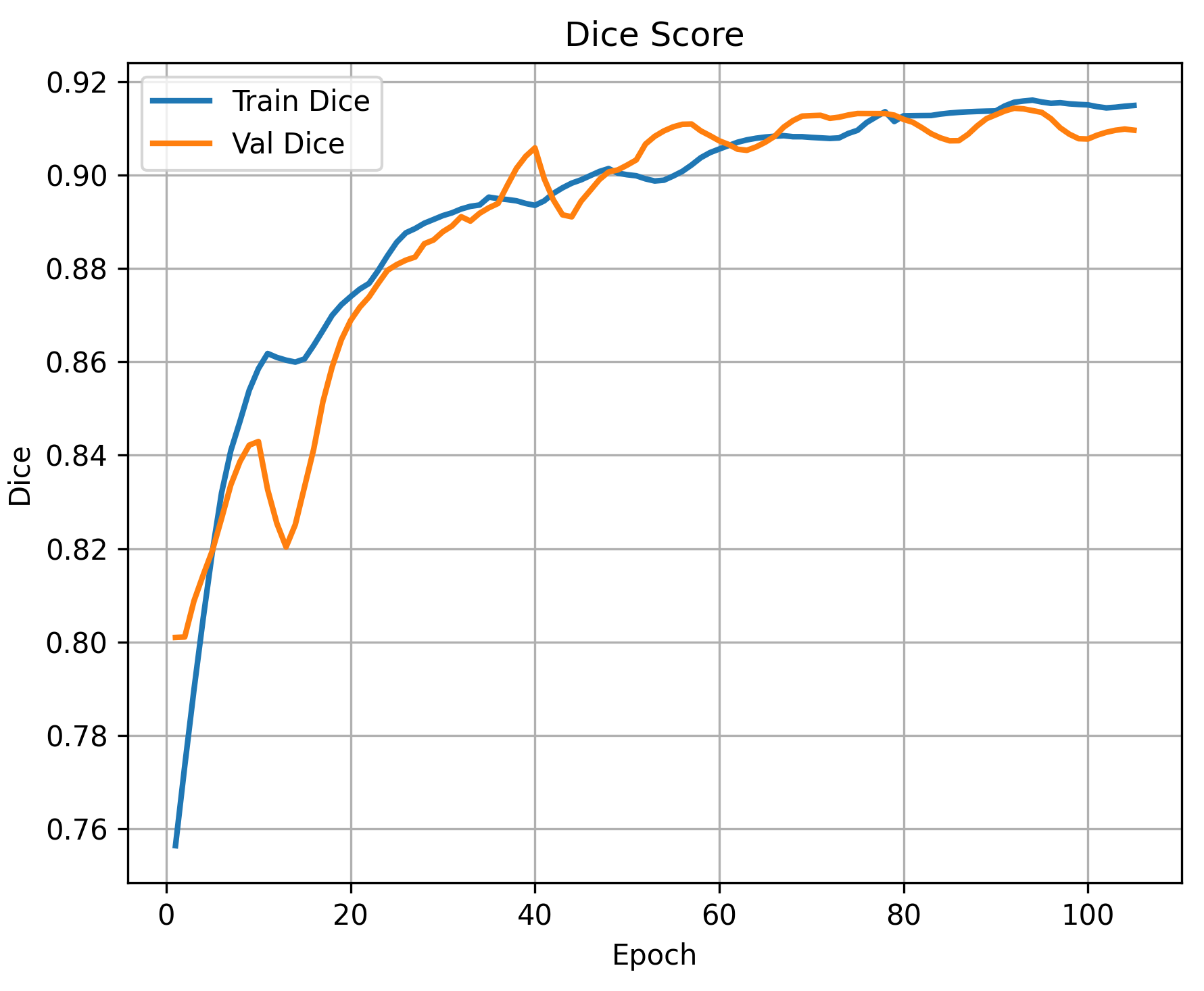}\\[1ex]
 
  \includegraphics[width=0.55\columnwidth]{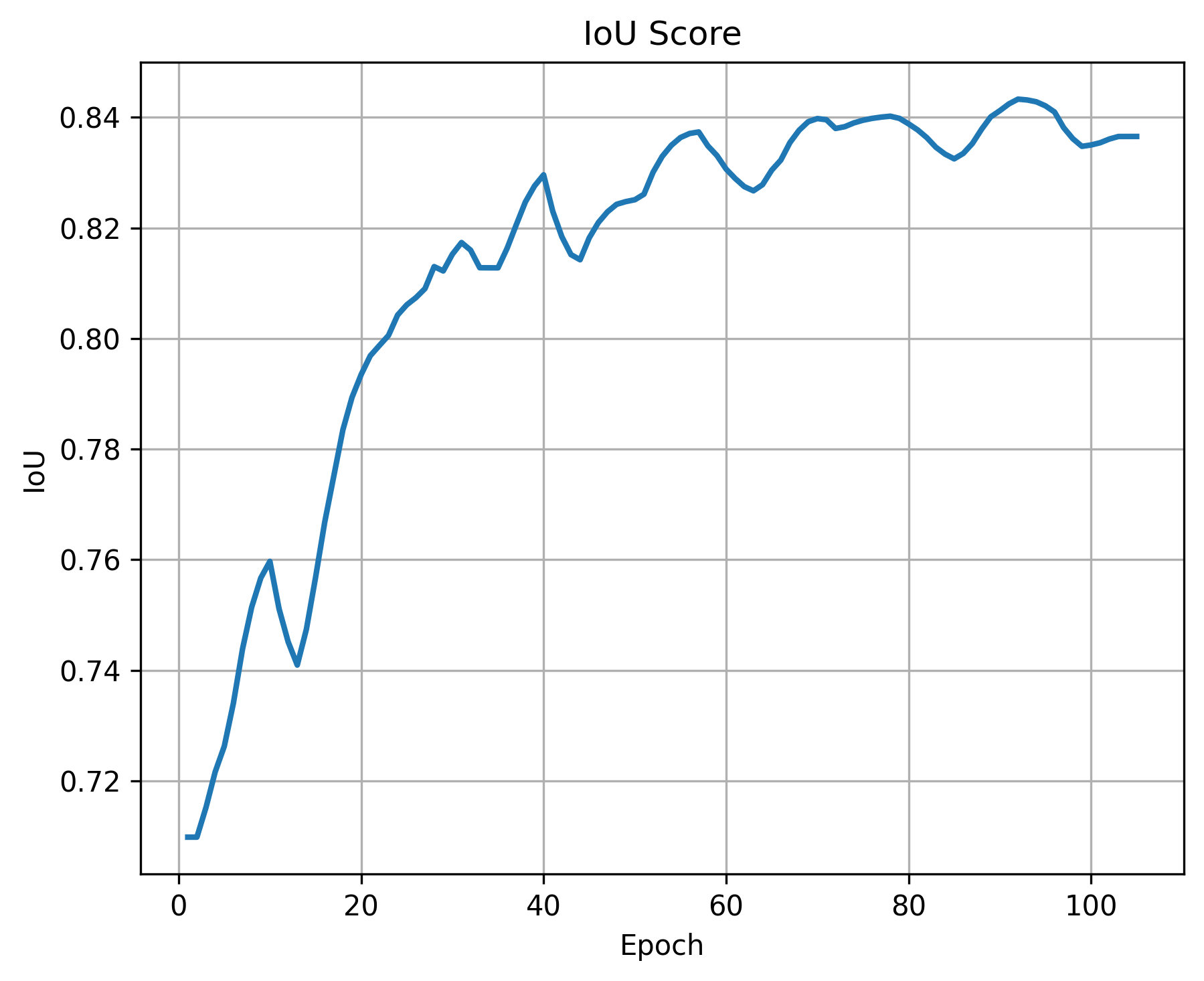}\\[1ex]
  \caption{Training loss, Dice score, and IoU score plotted over the course of training epochs.}
  \label{fig:three_metrics}
\end{figure}

Figure~\ref{fig:three_metrics} illustrates the training convergence characteristics of MedLiteNet. The training loss exhibits monotonic decrease with iterations, stabilizing after approximately 50 epochs, indicating effective optimization convergence.

Regarding performance metrics, both training and validation Dice coefficients demonstrate rapid improvement in the initial phase, ultimately converging to a high-performance range of 0.86-0.91. The validation IoU follows a similar trajectory, stabilizing above 0.80. Notably, the training and validation curves exhibit remarkable alignment, with minimal divergence maintained throughout the training process, indicating absence of overfitting. This consistency suggests robust generalization capability and the model's ability to learn transferable features.

Convergence speed analysis reveals that the model reaches performance saturation around epoch 50, with subsequent iterations showing only minor fluctuations in all metrics. This rapid convergence characteristic validates the effectiveness of the proposed architecture and facilitates potential rapid deployment in clinical settings.

\subsection{Limitations and Future Directions}

While MedLiteNet demonstrates excellent segmentation performance, analyzing its limitations provides valuable insights for future improvements.

Current Limitations: First, boundary prediction accuracy remains suboptimal in specific scenarios. When lesion boundaries are ambiguous or morphologically complex, the model may produce subtle under- or over-segmentation, failing to precisely delineate true contours. Second, low-contrast detection capability requires enhancement. When visual differences between lesions and surrounding healthy skin are minimal, the model may exhibit localized false negatives, particularly pronounced in regions with uneven illumination or skin tone variations. Third, robustness to external interference is insufficient. Non-lesion elements such as hair follicles and measurement rulers are occasionally misidentified, resulting in segmentation artifacts or discontinuities.

\begin{figure}[ht]
      \centering
      \includegraphics[width=0.8\linewidth]{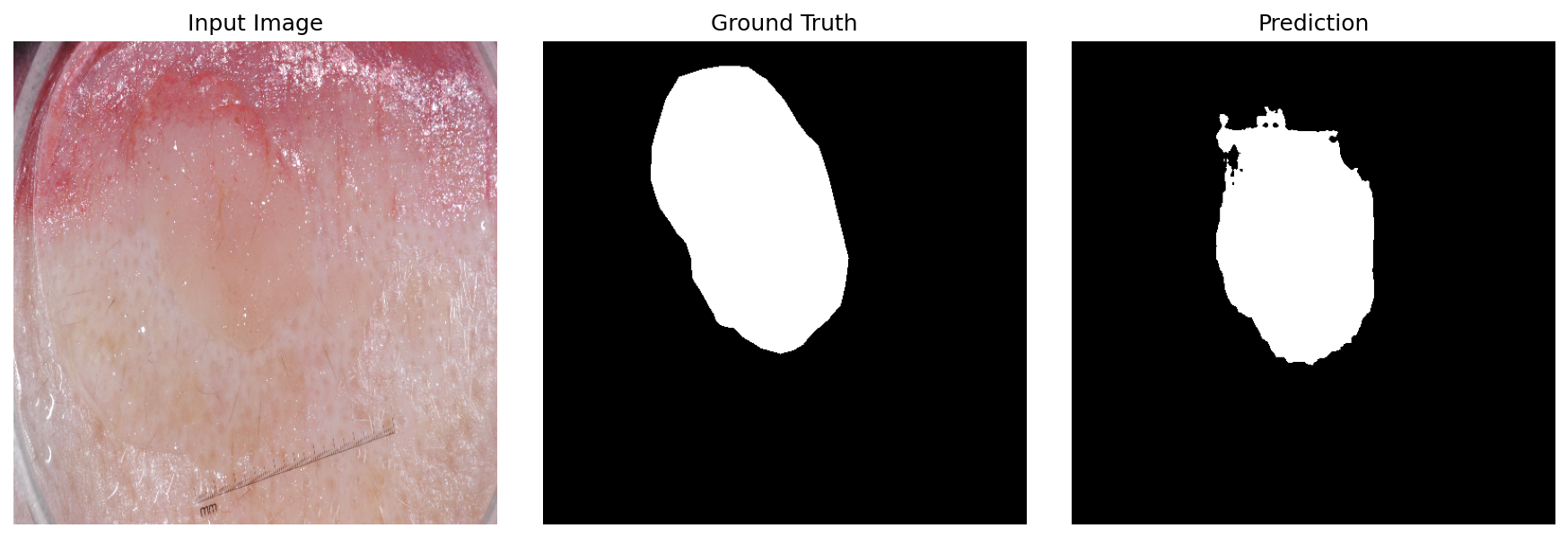}
      \caption{Example comparison: input image (left), human-annotated ground truth (middle), and MedLiteNet segmentation result at blurred lesion boundaries (right).}
      \label{fig:robustness_failure}
    \end{figure}

Improvement Strategies: Based on the above analysis, we propose the following enhancement directions:

\begin{itemize}
    \item Preprocessing optimization: Develop specialized image preprocessing modules to automatically detect and remove interference factors such as hair follicles and shadows, reducing their impact on segmentation accuracy.
    
    \item Hybrid approaches: Integrate traditional edge detection operators with deep learning features to enhance the model's capability in perceiving lesion boundaries, particularly for cases with ambiguous contours.
    
    \item Targeted data augmentation:Expand the representation of low-contrast lesion samples in the training set and design specific augmentation strategies to simulate various lighting conditions and skin tone variations.
    
    \item Architectural improvements:Introduce specialized attention mechanism variants designed to strengthen feature extraction capabilities in low-contrast regions, enabling better detection of subtle lesions.
\end{itemize}

These improvements are expected to further enhance MedLiteNet's performance in complex clinical scenarios, advancing automated skin lesion analysis toward higher accuracy and robustness.

\section{Conclusion}

This paper presents MedLiteNet, a lightweight CNN-Transformer hybrid model that achieves efficient and accurate medical image segmentation through the integration of multiple innovative modules. The proposed model incorporates four key architectural innovations that collectively address the challenges of balancing accuracy and computational efficiency in medical image analysis.

First, MedLiteNet employs an inverted residual encoder as the feature extraction backbone, effectively extracting rich multi-scale feature representations while maintaining the network's lightweight characteristics. Second, the designed Local-Global fusion block intelligently combines the advantages of convolutional neural networks in local detail extraction with Transformer's capability in global context modeling, fully leveraging the complementary nature of both architectures to enhance segmentation performance.

Third, the model introduces a boundary-aware attention mechanism that adaptively enhances focus on lesion boundary details, facilitating more refined and accurate target contour segmentation results. Additionally, the adaptive channel and spatial weight module performs fine-grained feature map recalibration through adaptive weight allocation in both channel and spatial dimensions, effectively highlighting critical information while suppressing redundant features, thereby enhancing overall feature representation capability.

Experimental results comprehensively validate MedLiteNet's excellent balance among segmentation accuracy, computational efficiency, and model compactness. On the ISIC 2018 skin lesion segmentation dataset, the model achieved approximately 0.91 Dice coefficient and 0.83 IoU metrics, with single-frame inference time of only about 1 millisecond and parameter count of merely 3.2M. These performance indicators significantly outperform the comprehensive performance of most existing methods, demonstrating the effectiveness of the proposed approach.

Comparative analysis with other real-time segmentation models in the literature shows that MedLiteNet's comprehensive Dice and IoU performance almost achieves the best segmentation accuracy, and also shows significant advantages in inference speed and model size. This perfect combination of high accuracy and high efficiency makes the model very suitable for deployment on resource-constrained mobile or edge devices, with significant clinical real-time application value and broad practical prospects.

Future work will explore the following research directions to continuously improve model performance: introducing multi-task learning mechanisms to achieve joint optimization of lesion classification and boundary detection, enhancing model adaptability to complex boundary morphologies and low-contrast lesions, and extending the model's generalization capability in multi-modal medical image segmentation tasks, thereby continuously improving segmentation performance and robustness.

\end{document}